\documentclass[conference]{IEEEtran}
\IEEEoverridecommandlockouts
\usepackage{cite}
\usepackage{amsmath,amssymb,amsfonts}
\usepackage{graphicx}
\usepackage{textcomp}
\usepackage{xcolor}
\usepackage{algpseudocode}

\def\BibTeX{{\rm B\kern-.05em{\sc i\kern-.025em b}\kern-.08em
    T\kern-.1667em\lower.7ex\hbox{E}\kern-.125emX}}
\begin{document}

\title{Deep Learning-Based Weather-Related Power Outage Prediction with Socio-Economic and Power Infrastructure Data}

\author{\IEEEauthorblockN{Xuesong Wang, Nina Fatehi, Caisheng Wang*\thanks{*Corresponding author.}, Masoud H. Nazari}
\\\small Department of Electrical and Computer Engineering, Wayne State University, Detroit, MI, USA
\\\small xswang@wayne.edu, nina.fatehi@wayne.edu, cwang@wayne.edu, masoud.nazari@wayne.edu}

\maketitle

\begin{abstract}

This paper presents a deep learning-based approach for hourly power outage probability prediction within census tracts encompassing a utility company's service territory. Two distinct deep learning models, conditional Multi-Layer Perceptron (MLP) and unconditional MLP, were developed to forecast power outage probabilities, leveraging a rich array of input features gathered from publicly available sources including weather data, weather station locations, power infrastructure maps, socio-economic and demographic statistics, and power outage records. Given a one-hour-ahead weather forecast, the models predict the power outage probability for each census tract, taking into account both the weather prediction and the location's characteristics. The deep learning models employed different loss functions to optimize prediction performance. Our experimental results underscore the significance of socio-economic factors in enhancing the accuracy of power outage predictions at the census tract level.

\end{abstract}


\section{Introduction}

Modern societies heavily rely on the power grid to supply continuous power to consumers. Power outages can cause serious and enormous costs. Extreme weather and natural disasters, such as hurricanes, wildfires, and floodings, are inevitable, which disturb the normal operation of power systems. According to \cite{R15}, 44\%-78\% of the reported power grid outages are weather-related, and the cost estimates from storm-related outages to the U.S. economy are between \$20 and \$55 billion annually. Therefore, rapid response to such events is crucial. The response includes preparing for and adapting to changes as well as quick recovery and restoration from major disruptions \cite{R7}. Therefore, there is a significant need to accurately forecast the extreme event impacts on the power grid so that utility crews can be dispatched during or even before the events and that long-term hardening approaches can be carried out in less resilient communities. 

Deep learning (DL) based models have also been proposed to predict power outages. A Recurrent Neural Network (RNN) is developed in \cite{R14} to predict outage duration by capturing temporal patterns in the outage data during extreme weather events. Long Short-Term Memory (LSTM) models are also developed to predict power load and power outage probability \cite{R18}. In \cite{R16}, authors compare the performances of different classification models for power outage prediction, e.g., Random Forest, Multi-Layer Perceptron (MLP), Support Vector Machine, K-Nearest Neighbor (KNN),  AdaBoost, and Quadratic Discriminant Analysis models. The authors of \cite{R19} combine Convolutional Neural Network (CNN) and LSTM to predict outages for 10 randomly selected counties across New York State. 

Recently, conditional models have attracted much attention in AI Generated Content (AIGC) \cite{R22, R23}. In conditional models, a condition is given to the model to modulate the base features. In the weather-related power outage prediction, weather data is the base input, and the characteristics of the location are the condition \cite{R8, R21}. Different local conditions, e.g., infrastructure age and soil conditions, should lead to various outage probabilities under the same weather condition.

The inputs to the models in the power outage prediction context usually include weather information, power system details, environmental conditions, and demographic data \cite{R8}. In \cite{R13}, wind speed and the distance of each component from the center of hurricane are used. A derived resiliency index for each component is utilized in \cite{R3}. To predict power load, historical power load data are utilized in \cite{R18}. To study the socio-economic impacts on electricity-vulnerable population groups, e.g., household income and racial demographics, authors of \cite{R17} designed an interactive tool to collect, analyze and predict power outages in New York City. In power outage prediction, various targets are employed, such as power outage occurrence (outage or non-outage) \cite{R16}, number of customers who lose power \cite{R8} and the ratio of power outage customers over the population \cite{R21}. Sparsity or unbalanced data is an issue in power outage prediction. To address it, authors of \cite{R20} propose a Majority Under-Sampling and Minority Over-Sampling strategy to balance the non-outage and outage samples. 

In this paper, we consider two deep learning models, an unconditional MLP and a conditional MLP, to predict the number of customers who lose power in an hourly resolution, normalized by the population of the census tracts. We show the effectiveness of the conditional model in power outage predictions. For the input data, we consider not only weather features, but also the characteristics of the communities (census tracts) in the DTE Energy service territory and show how socio-economic factors impact the power outage predictions. We also propose using cross-entropy loss function and assigning a large weight to the outage class to mitigate the class imbalance issue in the dataset and compare the performance of the weighted cross entropy loss and an exponential loss on both models. Therefore, our main contribution can be summarized as follows:

\begin{itemize}
    \item We have developed two deep learning models, an unconditional MLP and a conditional MLP, to predict power outage probability with census tract characteristics as the conditional input.
    \item We have compared the performance of the conditional MLP with unconditional MLP under two different loss functions: a weighted cross entropy loss and an exponential loss. 
    \item We have investigated the impact of socio-economic factors as well as power infrastructure features on power outage frequencies. The results show that these factors effectively reduce the error rate in the proposed models. 
\end{itemize}

The remaining sections are arranged as follows: In Section II, the data collection and preprocessing are discussed. This is followed by the introduction of the proposed power outage prediction method in section III. Section IV presents the performance comparison between the conditional and unconditional models and various input factors. In Section V, the conclusion is summarized.

\section{Data Collection and Preprocessing}

\subsection{Outage Data}

\begin{figure}[tbp]
\centerline{\includegraphics[scale=0.35]{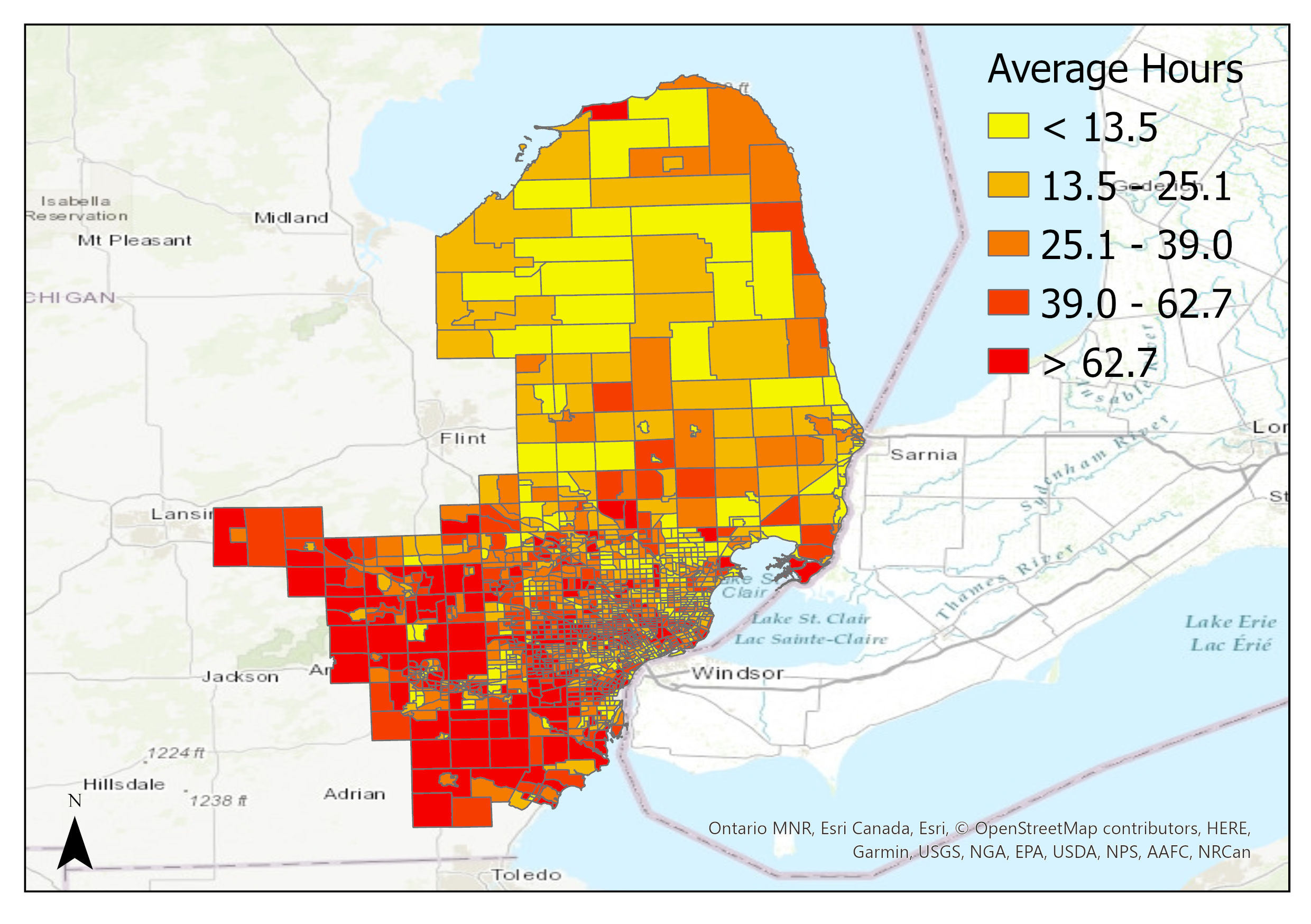}}
\caption{The average power outage duration in hours per capita for each census tract.}
\label{census_out_dist}
\end{figure}

The area under investigation encompasses the DTE's service territory in Michigan. Through DTE's Outage Management System interface, outage data from Feb. 27, 2023 to Sep. 30, 2023 were collected every 15 minutes. For each power outage event, we take its first occurrence as the start time and the last occurrence as the end time. As the spatial coverage and number of customers may change during the event period, we take the largest spatial coverage and the maximum number of customers during its period as its coverage and affected customers, respectively. To align the power outage events to the census tract level, a pairwise intersection operation is executed in ArcGIS Pro, distributing the affected customers proportionally. The average power outage duration in hours per capita for each census tract is shown in Fig.~\ref{census_out_dist}. After removing all the power outage events affecting less than or equal to 1 customer, we count the number of customers that lost power in each hour in the studied period. In total, 1,102 census tracts are selected, and for each census tract, 5,184 hourly power outage probabilities are calculated. Note that outages usually occur after storms or high wind gusts, resulting in the sparsity of outages in the dataset, i.e., on average 26 out of 5184 hours are outage hours for all census tracts.

\begin{figure}[tbp]
\centerline{\includegraphics[scale=0.34]{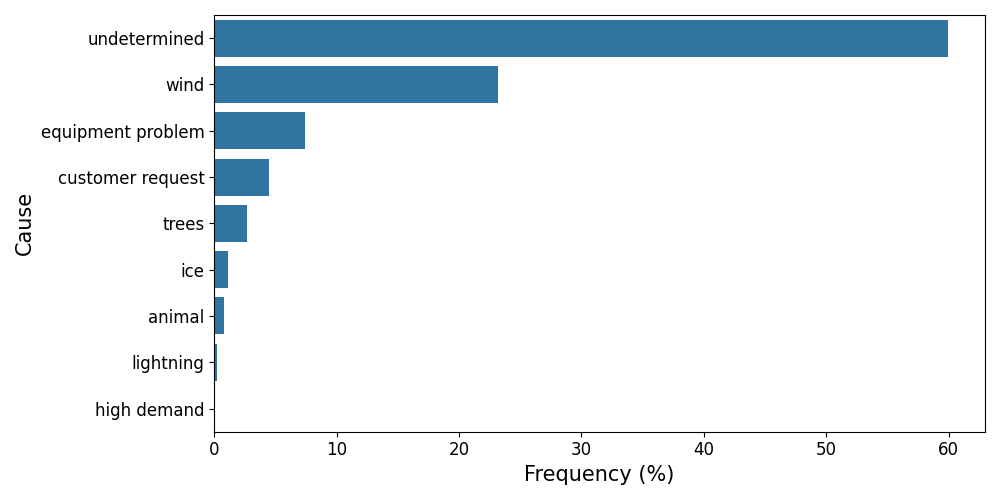}}
\caption{Histogram of the causes of the power outage events.}
\label{out_causes}
\end{figure}

In cases where the number of affected customers surpasses the population count of the respective census tract, we assign a probability value of 1. Note that while certain power outage events may not be solely attributed to weather conditions, we retain them in the dataset for two primary reasons. Firstly, the majority of identified event causes are weather-related, such as wind or ice impacts. Secondly, a substantial number of events lack a definitive cause within the records, as illustrated in Fig.~\ref{out_causes}. We consider the variability present in the power outage data as a form of data augmentation and a regularization mechanism for our model.

\subsection{Weather and Weather Station Data}

The historical hourly weather data are collected from Automated Surface Observing System (ASOS).\footnote{Downloaded from https://mesonet.agron.iastate.edu/ASOS/} We select 10 features from hourly observations, including air temperature, relative humidity, pressure altimeter, wind speed and direction, etc. Regarding temperature, humidity, and air pressure data, missing values are addressed through temporal interpolation. As for the remaining features, any missing values are filled with zeros. In cases where observations are missing, we employ the average of the available data from other weather stations. Subsequently, the data is normalized by mean and standard deviation.

\subsection{Power Infrastructure Data}

\begin{figure}[bp]
\centerline{\includegraphics[scale=0.22]{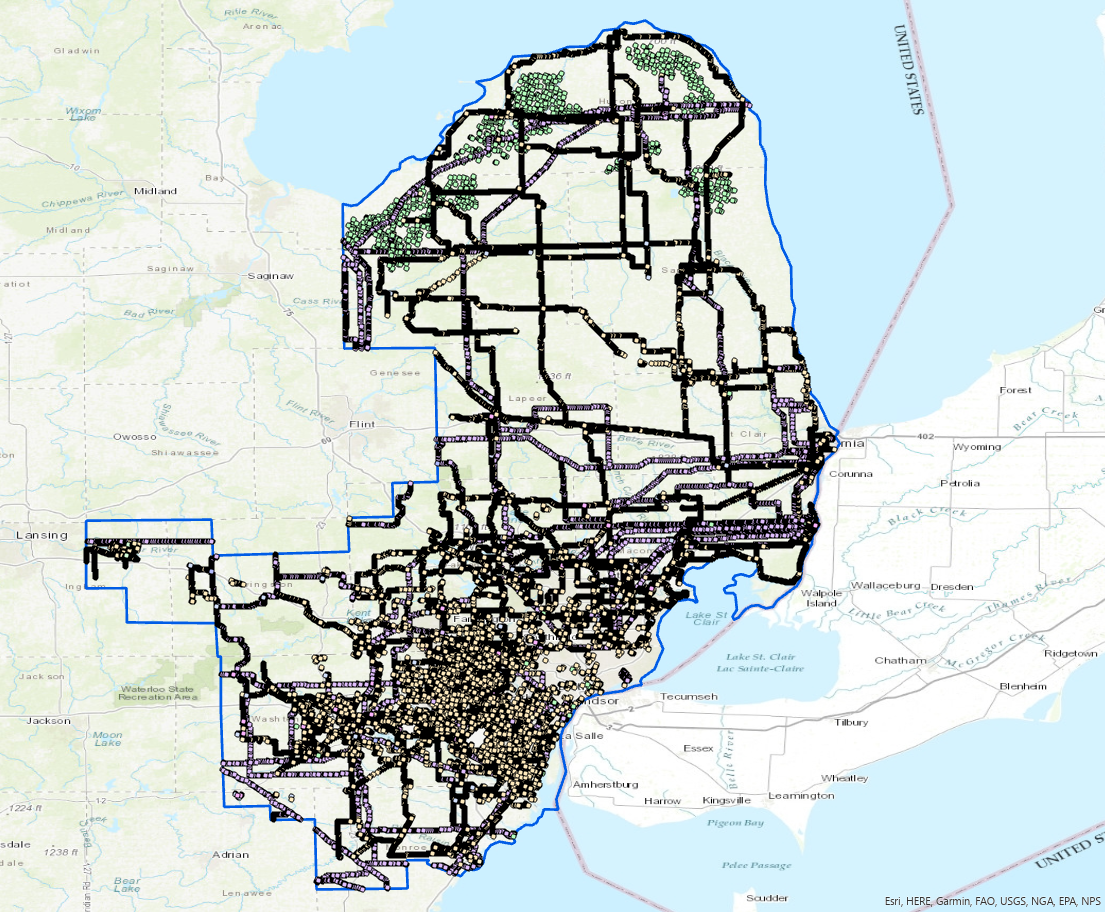}}
\caption{Power infrastructure components in the studied area.}
\label{power_infra}
\end{figure}

The power infrastructure data are collected from Open Street Map dataset \cite{R24}, as shown in Fig.~\ref{power_infra}. All nodes in the studied area that have attribute "power" are collected. The types of nodes include compensator, generator, insulator, line, pole, portal, substation, switch, terminal, tower, and transformer. The number of each feature is counted for census tracts using ArcGIS Pro. Note that this dataset may not ensure data completeness. Nonetheless, we consider the inherent noise within the data as a form of data augmentation and a regularization technique for our model. Since the detailed topology of the power distribution network is usually not publicly available, the Open Street Map data \cite{R24} can serve as a good approximation of the true topology for research purposes.

\subsection{Socio-economic and Demographic Data}

Even when exposed to identical weather conditions, census tracts with varying levels of resilience are likely to encounter distinct degrees of power outages. For each census tract, we collect the following data from the U.S. Census Bureau:

\begin{itemize}
    \item Total population in 2020
    \item Household income distribution estimated in 2021
    \item Distribution of year structure built estimated in 2021
\end{itemize}

The household income and year structure built data include both estimations and margins of error. Prior to normalizing the distribution with Softmax, we employ a data augmentation technique during model training. This augmentation involves generating a uniform distribution based on the mean and margin of error for each data bin.

\section{Proposed Power Outage Prediction Method}

We design two model structures, i.e., unconditional and conditional models. For each model, we utilize two loss functions, i.e., exponential loss and cross entropy loss. In the following subsections, we elaborate the proposed models.

\subsection{Conditional MLP Structure}

\begin{figure}[bp]
\centerline{\includegraphics[scale=0.2]{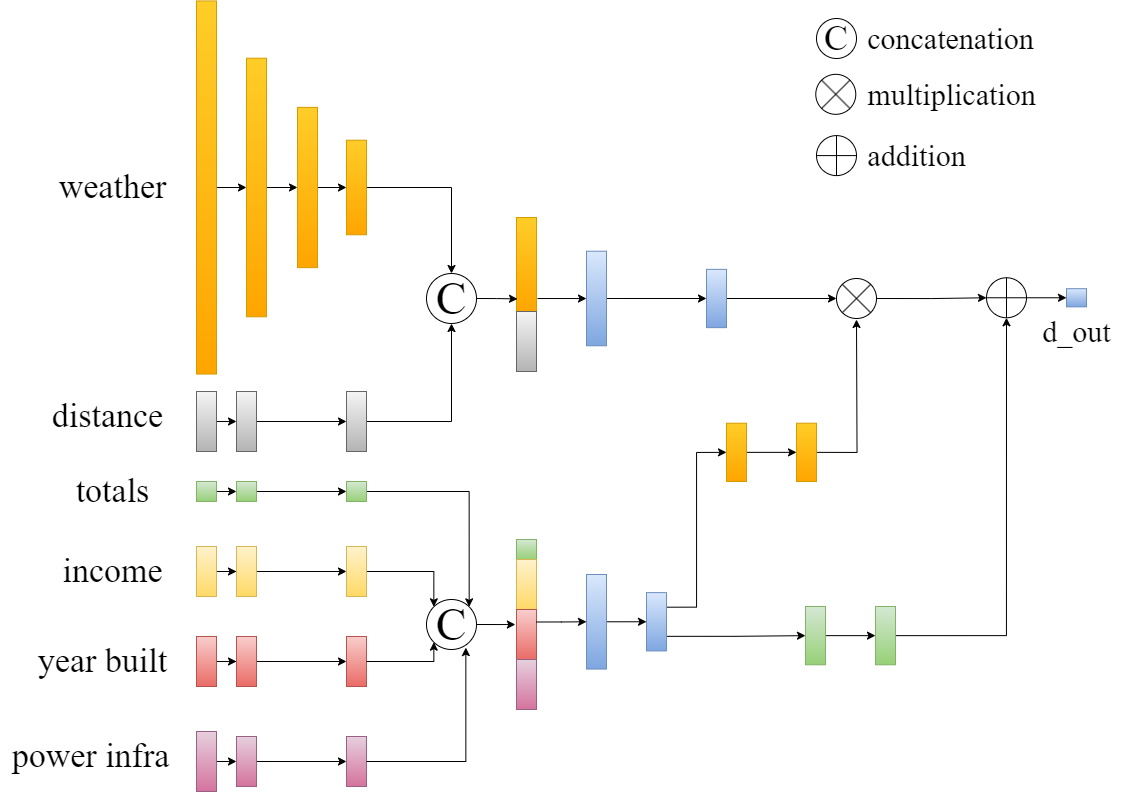}}
\caption{Structure of the conditional model, d\_out=1 for the exponential loss and d\_out=2 for the weighed cross entropy loss}
\label{conditional_model}
\end{figure}

The conditional model comprises two primary branches, as illustrated in Fig.~\ref{conditional_model}. The first branch takes weather features and the distances between weather stations and the census tract as the input. These inputs are then processed through several linear and non-linear layers. The second branch receives four types of inputs: 1) household income distribution, 2) year structure built distribution, 3) power infrastructure distribution, and 4) totals (including the total population, total households, total houses,  and total power infrastructure components). These inputs are then processed through a series of linear and non-linear layers. The output vector generated by the second branch undergoes a transformation involving two sets of layers, resulting in a scale vector and a bias vector. These two vectors are then applied to the output of the first branch using \eqref{condition}, and the resulting output is further processed by the final output layer. This design is driven by the concept that, when confronted with identical weather conditions, the model should adapt the influence of weather on the power outage probability for a given census tract based on its infrastructure conditions. These conditions can be characterized by features such as the distribution of structure built years and the attributes of power infrastructure.

\begin{equation}
F_{out} = F_{in} * scale + bias
\label{condition}
\end{equation}

\subsection{Unconditional MLP Structure}

\begin{figure}[tbp]
\centerline{\includegraphics[scale=0.2]{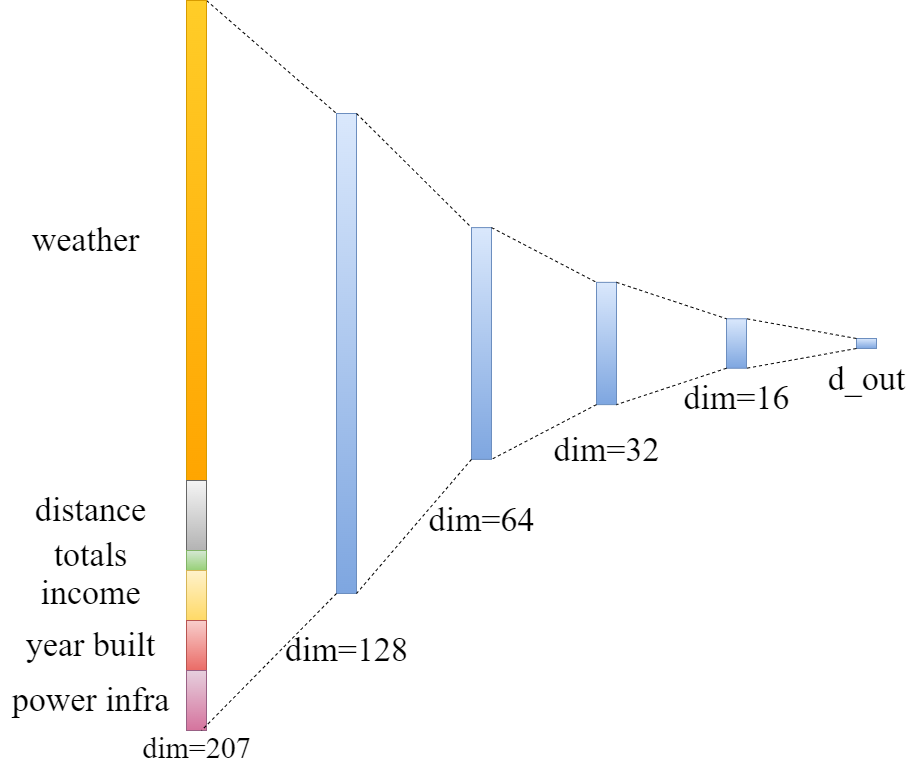}}
\caption{Structure of the unconditional model, d\_out=1 for the exponential loss and d\_out=2 for the weighed cross entropy loss}
\label{unconditional_model}
\end{figure}

The unconditional model concatenates all input features into a 1-D vector, subsequently passing it through a series of alternating linear and non-linear layers. These layers exhibit a monotonically decreasing number of neurons in each layer. The structure is shown in Fig.~\ref{unconditional_model}.

\subsection{Power Outage Probability Prediction}

The prediction target is the probability of power outage where the values are between 0 and 1. One can naturally think of using cross entropy loss. Because of the imbalance between outage and non-outage classes, we put a large weight on the outage class. The outputs of both models go through a Softmax layer and the cross entropy losses are calculated by \eqref{cross_entropy}, where $N$ is the number of samples in the training set, $P_{gt,0}$ and $P_{pred,0}$ are the probability of non-outage from ground truth and model prediction, respectively. Also, $P_{gt,1}$ and $P_{pred,1}$ are the probability of outage from ground truth and model prediction, and $w$ is the weight for the outage class ($w=500$ in our case).

\begin{equation}
L = -\frac{1}{N}\sum{[P_{gt,0}\log{P_{pred,0}}+w * P_{gt,1}\log{P_{pred,1}}]}
\label{cross_entropy}
\end{equation}

The second loss function employed is an exponential loss function, as depicted in \eqref{exponential_loss}. Here, $P_{gt}$ and $P_{pred}$ represent the probability of power outage in the ground truth and the model prediction, respectively. The parameter $\beta$ serves as a scaling factor to penalize substantial disparities between the prediction and ground truth (in our instance, $\beta$ is set to 20).

\begin{equation}
L = \frac{1}{N}\sum{e^{|P_{gt} - P_{pred}| * \beta}}
\label{exponential_loss}
\end{equation}

\begin{figure}[tbp]
\centerline{\includegraphics[scale=0.42]{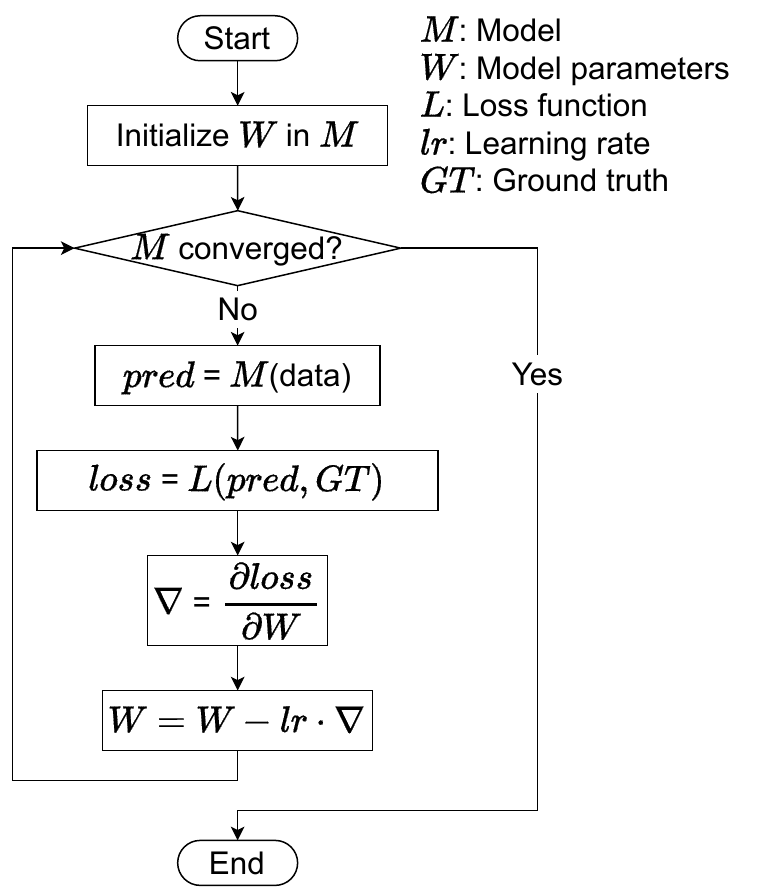}}
\caption{Model training using back propagation.}
\label{backprop}
\end{figure}

Both models are trained using back propagation as shown in Fig.~\ref{backprop}.

\subsection{Evaluation Metrics}

Because the outputs of Sigmoid and Softmax layers cannot reach 0, model outputs under 0.05 are set to 0 before evaluation. We use Mean Absolute Error (MAE) and Root Mean Squared Error (RMSE) to evaluate the model's performance, as shown in \eqref{mae} and \eqref{rmse}.

\begin{equation}
MAE = \frac{1}{N}\sum{|P_{gt} - P_{pred}|}\label{mae}
\end{equation}

\begin{equation}
RMSE = \sqrt{\frac{1}{N}\sum{|P_{gt} - P_{pred}|^2}}\label{rmse}
\end{equation}

\begin{table}[tbp]
\caption{Evaluation Metrics of Different Models and Loss Functions}
\begin{center}
\begin{tabular}{|c|c|c|c|}
\hline
Model & Loss Function & MAE & RMSE\\
\hline
unconditional & exponential & 0.01346(6.8e-3) & 0.04773(1.5e-2)\\
\hline
conditional & exponential & 0.0189(6.5e-3) & 0.04824(5.2e-3)\\
\hline
unconditional & cross entropy & 0.03243(7.1e-4) & 0.1179(2.2e-3)\\
\hline
conditional & cross entropy & 0.0281(1.3e-3) & 0.1131(4.4e-3)\\
\hline
\end{tabular}
\label{metrics}
\end{center}
\end{table}

\section{Experiment Results}

We partitioned the census tracts in the study area into three subsets: 72\% for the training set, 8\% for the validation set, and 20\% for the test set. Our evaluation metrics are reported based on the test sets. The results were averaged over three separate runs, showcasing both the mean and standard deviations. As observed in Table \ref{metrics}, when the exponential loss function is used, the unconditional model exhibits lower error, whereas with the cross-entropy loss, the conditional model demonstrates superior performance. The variation in performance between the two models may stem from the choice of hyper-parameters, such as the selection of the parameter $w$ in the cross-entropy loss and $\beta$ in the exponential loss.

\begin{figure}[tbp]
    \centering
    \includegraphics[scale=0.32]{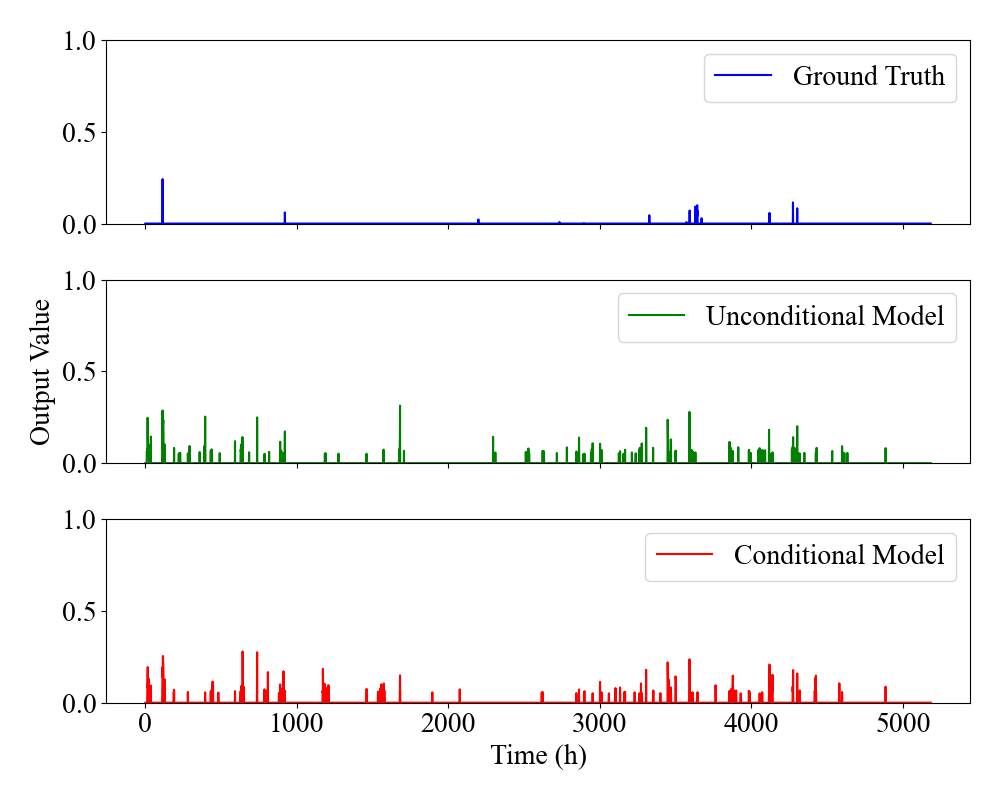}
    \caption{Predictions of both models using exponential loss for one of the census tracts.}
    \label{model_output}
\end{figure}

Predictions of both models using exponential loss for one of the census tracts are shown in Fig.~\ref{model_output}. As illustrated in Fig.~\ref{model_output}, false positive is the main source of error due to the data imbalance. A two-stage model might reduce the false positive prediction by first classifying the outage status (outage or non-outage) then estimating the outage customer number. Moreover, a Majority Under-Sampling and Minority Over-Sampling strategy can be utilized to overcome the class imbalance problem \cite{R20}.

\begin{table*}
\caption{Ablation Study of Different Input Factors}
\begin{center}
\begin{tabular}{|c|c|c|c|c|c|c|c|c|c|}
\hline
\multicolumn{6}{|c|}{Input Factors} & \multicolumn{2}{|c|}{Exponential} & \multicolumn{2}{|c|}{Cross Entropy} \\\hline 
Weather & Distance & Totals & Income & Year Built & Power Infra & MAE & RMSE & MAE & RMSE \\\hline
\checkmark & & & & & & 0.03547(7.3e-3) & 0.0703(4.5e-3) & 0.04901(3.5e-3) & 0.127(5.6e-3) \\\hline
\checkmark & \checkmark & & & & & 0.01916(1.0e-2) & 0.05018(1.2e-2) & 0.03652(1.6e-3) & 0.1231(3.8e-3) \\\hline
\checkmark & \checkmark & \checkmark & & & & 0.01015(4.3e-3) & 0.03777(8.7e-3) & 0.03662(3.1e-3) & 0.1239(5.8e-3) \\\hline
\checkmark & \checkmark & \checkmark & \checkmark & & & 0.01376(4.4e-3) & 0.04874(1.0e-2) & 0.0355(2.5e-3) & 0.1233(5.4e-3) \\\hline
\checkmark & \checkmark & \checkmark & \checkmark & \checkmark & & 0.01413(6.5e-3) & 0.04848(1.0e-2) & 0.03235(2.6e-3) & 0.1178(4.6e-3) \\\hline
\checkmark & \checkmark & \checkmark & \checkmark & \checkmark & \checkmark & 0.01346(6.8e-3) & 0.04773(1.5e-2) & 0.03243(7.1e-4) & 0.1179(2.2e-3) \\\hline
\end{tabular}
\label{ablation}
\end{center}
\end{table*}

To investigate the impacts of different factors on the model performance, we conducted an ablation study on the unconditional model. We gradually add more features into the model input and see the performance variations of the model. The mean and standard deviation of three runs are reported. The results are shown in Table \ref{ablation}. The inclusion of additional features leads to a reduction in prediction errors. The variability in results for models employing the exponential loss could potentially be attributed to noise in the household income, year structure built, and power infrastructure distribution data. 

\section{Conclusion}

This paper devised both unconditional and conditional models to predict power outage probabilities at the census tract level. To mitigate unbalanced data issue, two distinct loss functions were implemented: exponential loss and weighted cross entropy loss. The results showed that using the exponential loss results in 5\% less MAE in the unconditional model than the conditional model; however, the weighted cross entropy loss showed 0.4\% less MAE in the conditional model than the unconditional model. Furthermore, the ablation study, conducted to investigate the impact of various input features on the model performance, showed that incorporating socio-economic factors and power infrastructure features reduced the model's prediction error and, therefore, had a positive impact on the power outage probability prediction.

Future research avenues may involve further feature integration into the model, exploring the application of LSTM to leverage historical power outage patterns, or transitioning to a broader city or county-level analysis as a means to partially address the data imbalance issue encountered at the finer-grained census tract level.

\section*{Acknowledgment}

We express our gratitude to Mark Furland for generously supplying the historical DTE outages data. 
 
This material is based upon work supported by the Department of Energy, Solar Energy Technologies Office (SETO) Renewables Advancing Community Energy Resilience (RACER) program under Award Number DE-EE0010413. Any opinions, findings, conclusions, or recommendations expressed in this material are those of the authors and do not necessarily reflect the views of the Department of Energy.

\vspace{12pt}

\bibliographystyle{ieeetr}
\bibliography{references}

\begin{thebibliography}{10}

\bibitem{R15}
R.~J. Campbell and S.~Lowry, ``Weather-related power outages and electric system resiliency,'' tech. rep., Congressional Research Service, Library of Congress Washington, DC, 2021.

\bibitem{R7}
M.~M. Hosseini and M.~Parvania, ``Artificial intelligence for resilience enhancement of power distribution systems,'' {\em The Electricity Journal}, vol.~34, no.~1, p.~106880, 2021.

\bibitem{R14}
A.~Jaech, B.~Zhang, M.~Ostendorf, and D.~S. Kirschen, ``Real-time prediction of the duration of distribution system outages,'' {\em IEEE Transactions on Power Systems}, vol.~34, no.~1, pp.~773--781, 2018.

\bibitem{R18}
M.~Abaas, R.~A. Lee, and P.~Singh, ``Long short-term memory customer-centric power outage prediction models for weather-related power outages,'' in {\em 2022 IEEE Green Energy and Smart System Systems (IGESSC)}, pp.~1--6, 2022.

\bibitem{R16}
Y.~Kor, M.~Z. Reformat, and P.~Musilek, ``Predicting weather-related power outages in distribution grid,'' in {\em 2020 IEEE Power and Energy Society General Meeting (PESGM)}, pp.~1--5, 2020.

\bibitem{R19}
K.~Udeh, D.~W. Wanik, D.~Cerrai, D.~Aguiar, and E.~Anagnostou, ``Autoregressive modeling of utility customer outages with deep neural networks,'' in {\em 2022 IEEE 12th Annual Computing and Communication Workshop and Conference (CCWC)}, pp.~0406--0414, 2022.

\bibitem{R22}
L.~Zhang, A.~Rao, and M.~Agrawala, ``Adding conditional control to text-to-image diffusion models,'' in {\em Proceedings of the IEEE/CVF International Conference on Computer Vision}, pp.~3836--3847, 2023.

\bibitem{R23}
T.~Karras, S.~Laine, and T.~Aila, ``A style-based generator architecture for generative adversarial networks,'' in {\em Proceedings of the IEEE/CVF conference on computer vision and pattern recognition}, pp.~4401--4410, 2019.

\bibitem{R8}
P.~Arora and L.~Ceferino, ``Probabilistic and machine learning methods for uncertainty quantification in power outage prediction due to extreme events,'' {\em EGUsphere}, pp.~1--29, 2022.

\bibitem{R21}
D.~B. McRoberts, S.~M. Quiring, and S.~D. Guikema, ``Improving hurricane power outage prediction models through the inclusion of local environmental factors,'' {\em Risk analysis}, vol.~38, no.~12, pp.~2722--2737, 2018.

\bibitem{R13}
R.~Eskandarpour and A.~Khodaei, ``Machine learning based power grid outage prediction in response to extreme events,'' {\em IEEE Transactions on Power Systems}, vol.~32, no.~4, pp.~3315--3316, 2016.

\bibitem{R3}
A.~Imteaj, M.~H. Amini, and J.~Mohammadi, ``Leveraging decentralized artificial intelligence to enhance resilience of energy networks,'' in {\em 2020 IEEE Power and Energy Society General Meeting (PESGM)}, pp.~1--5, IEEE, 2020.

\bibitem{R17}
S.~Eckstrom, G.~Murphy, E.~Ye, S.~Acharya, R.~Mieth, and Y.~Dvorkin, ``Outing power outages: Real-time and predictive socio-demographic analytics for new york city,'' in {\em 2022 IEEE Power and Energy Society General Meeting (PESGM)}, pp.~1--5, 2022.

\bibitem{R20}
A.~Bahrami, M.~Shahidehpour, S.~Pandey, W.~Nation, K.~DSouza, and H.~Zheng, ``Machine learning application to extreme weather power outage forecasting in distribution networks using a majority under-sampling and minority over-sampling strategy,'' in {\em 2023 IEEE Power and Energy Society General Meeting (PESGM)}, pp.~1--6, 2023.

\bibitem{R24}
{OpenStreetMap contributors}, ``{Planet dump retrieved from https://planet.osm.org},'' 2023.
\newblock Accessed on Oct. 24, 2023.

\end{thebibliography}
\end{document}